\pdfoutput=1

\documentclass[11pt]{article}

\usepackage[preprint]{acl}

\usepackage{times}
\usepackage{latexsym}

\usepackage[T1]{fontenc}

\usepackage[utf8]{inputenc}

\usepackage{microtype}

\usepackage{inconsolata}

\usepackage{graphicx}
\usepackage{booktabs}
\usepackage{amsmath}
\usepackage{booktabs} 
\usepackage{multirow} 
\usepackage{makecell}
\usepackage[most]{tcolorbox}
\newcounter{promptbox}
\usepackage{float}

\title{THINK-Bench: Evaluating Thinking Efficiency and Chain-of-Thought Quality of Large Reasoning Models}

\author{
 \textbf{Zhiyuan Li\textsuperscript{1}},
 \textbf{Yi Chang\textsuperscript{1,2,3}},
 \textbf{Yuan Wu\textsuperscript{1}\thanks{Corresponding author}}
\\
 \textsuperscript{1} School of Artificial intelligence, JiLin University
 \\
 \textsuperscript{2} Engineering Research Center of Knowledge-Driven Human-Machine Intelligence, JiLin University
 \\
 \textsuperscript{3} International Center of Future Science, JiLin University
\\
 \small{
   \textbf{Correspondence:} \href{mailto:yuanwu@jlu.edu.cn}{yuanwu@jlu.edu.cn}
 }
 \\
 \small{
 \textbf{Project Page:} \href{https://think-bench.github.io/}{https://think-bench.github.io/}
 }
}

\begin{document}
\maketitle

\begin{abstract}
Large reasoning models (LRMs) have achieved impressive performance in complex tasks, often outperforming conventional large language models (LLMs). However, the prevalent issue of overthinking severely limits their computational efficiency. Overthinking occurs when models generate excessive and redundant tokens that contribute little to accurate outcomes, especially in simple tasks, resulting in a significant waste of computational resources. To systematically investigate this issue, we introduce Think-Bench, a benchmark designed to evaluate the reasoning efficiency of LRMs. We also propose novel efficiency metrics and conduct a comprehensive evaluation of various LRMs across multiple dimensions, including the reasoning process, outcome quality, and chain-of-thought (CoT) characteristics. Our analysis reveals that most LRMs exhibit overthinking in handling easy questions, generating unnecessarily lengthy reasoning chains. While many LRMs demonstrate high CoT quality, several suffer from low efficiency. We hope that Think-Bench can serve as a robust foundation for advancing research into LRMs.
\end{abstract}
\begin{figure}
    \centering
    \includegraphics[width=\linewidth]{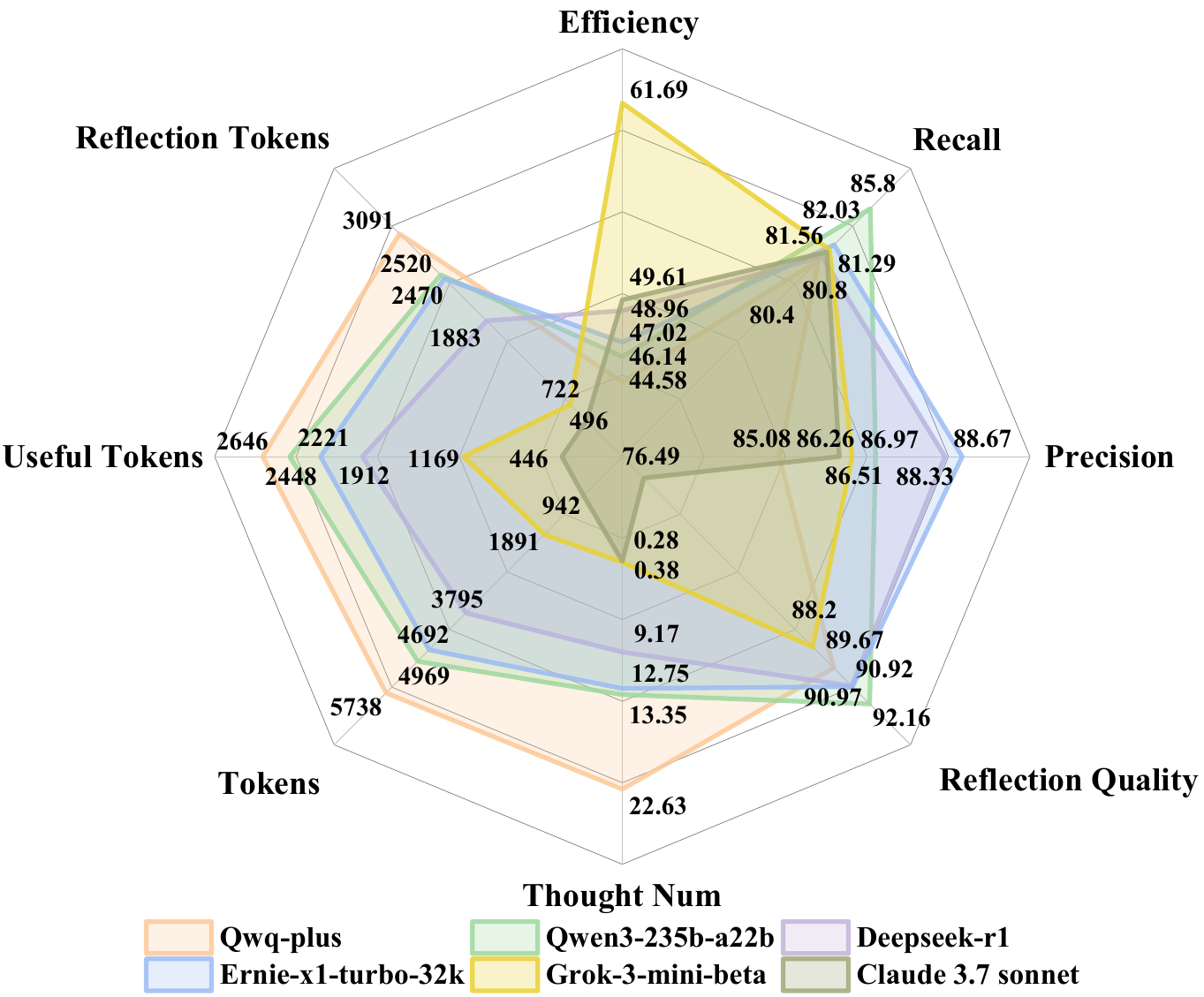}
    \caption{\textbf{The performance of various LRMs on Think-Bench.} The results suggest that these prominent LRMs face a challenge of overthinking.}
    \label{fig:enter-label}
\end{figure}

\section{Introduction}
In recent years, with the rapid advancement of artificial intelligence, LLMs have achieved remarkable success in the field of natural language processing (NLP), particularly excelling in tasks such as text generation and question answering~\citep{grattafiori2024llama, guo2025deepseek, yang2024qwen2}. However, the reasoning capabilities of these models in solving multidisciplinary problems still face significant challenges, such as insufficient integration of cross-disciplinary knowledge and weak logical chain reasoning ability~\citep{wang2024exploring,chen2025towards}. To gain a deeper understanding and effectively enhance the efficiency and accuracy of the reasoning capabilities of LLMs, constructing high-quality multidisciplinary datasets and conducting systematic evaluations has become critically important~\citep{chang2024survey,xia2024language}.

\begin{figure*}[!htbp]
    \centering
    \includegraphics[width=\linewidth]{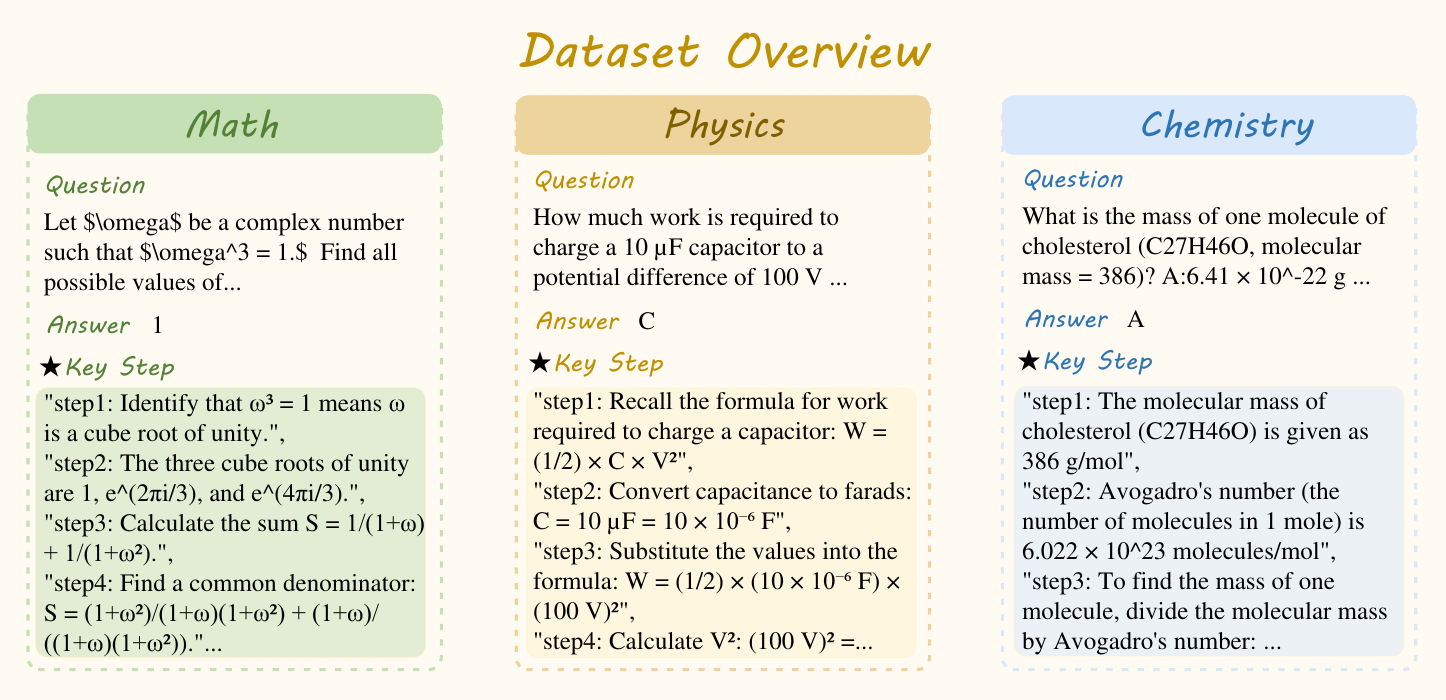}
    \caption{\textbf{Overview of Think-Bench.} Our benchmark contains a comprehensive efficiency evaluation framework with curated datasets across three categories.}
    \label{fig:datasetOverview}
\end{figure*}

Recent research have introduced several high-quality multidisciplinary benchmarks, such as MMLU~\citep{hendrycks2020measuring} and GPQA~\citep{rein2024gpqa}, to evaluate model performance across diverse knowledge domains. However, these datasets predominantly rely on the correctness of the final answer as the sole evaluation metric. Such outcome-oriented evaluation protocols are not directly applicable to the assessment of large reasoning models (LRMs), as they neglect the accuracy and logical coherence of intermediate reasoning steps during answer generation~\citep{jiang2025mme,zheng2024processbench}. In reality, for complex tasks spanning logical reasoning, mathematical problem-solving, and multi-step decision-making, the quality of intermediate CoT processes not only determines final-answer accuracy but also serves as a critical indicator of LRMs' intrinsic reasoning capabilities and operational reliability. Existing evaluation paradigms that ignore CoT quality risk masking systemic deficiencies, such as erroneous premises, logical fallacies, or context misinterpretations, which may propagate through the reasoning chain. Consequently, establishing a holistic evaluation framework that quantifies both intermediate-step correctness and final-output accuracy emerges as a pivotal yet understudied research direction.

Recently, several studies have begun to explore the evaluation of the correctness of CoTs. Notable works such as MME-CoT~\citep{jiang2025mme} establish a comprehensive benchmark that systematically examines CoTs across three fundamental dimensions: reasoning quality, robustness, and efficiency. Meanwhile, MiCEval~\citep{zhou2024miceval} conducts a fine-grained evaluation of Multimodal Chains of Thought (MCoT), focusing on aspects such as the accuracy of image descriptions and the correctness, relevance, and informativeness of reasoning steps. Collectively, these benchmarks represent a significant shift from solely outcome-oriented evaluations to more process-aware assessments that effectively capture the internal reasoning dynamics of large models.

Despite these advances, existing benchmarks remain limited by their lack of a systematic framework for evaluating both the efficiency and reliability of reasoning processes in LRMs.
This challenge is particularly prominent in  multidisciplinary problem-solving scenarios~\citep{wang2025thoughts}. Prior studies have demonstrated that reasoning models structurally similar to OpenAI-o1 often allocate computational resources inefficiently when handling relatively simple problems, while achieving only negligible performance improvements in final outcomes~\citep{chen2024not, guo2025deepseek}. These findings suggest that LRMs may suffer from suboptimal resource distribution during inference, consequently constraining their overall performance optimization.

To bridge this critical research gap, we introduce \textbf{Think-Bench}, a multidisciplinary dataset specifically designed for comprehensively evaluating the reasoning efficiency and accuracy of LRMs. Our benchmark features meticulously annotated key reasoning steps for each problem instance, enabling granular analysis of model-generated reasoning processes. Additionally, we propose an evaluation protocol that can measure the efficiency and rationality of the reasoning process by analysing the specific behaviours exhibited by LRMs during their thinking process.
\section{Dataset Curation}
\subsection{Data Overview}

\begin{figure*}[htbp]
    \centering
    \begin{minipage}{0.55\textwidth}
        \centering
        \includegraphics[width=\linewidth]{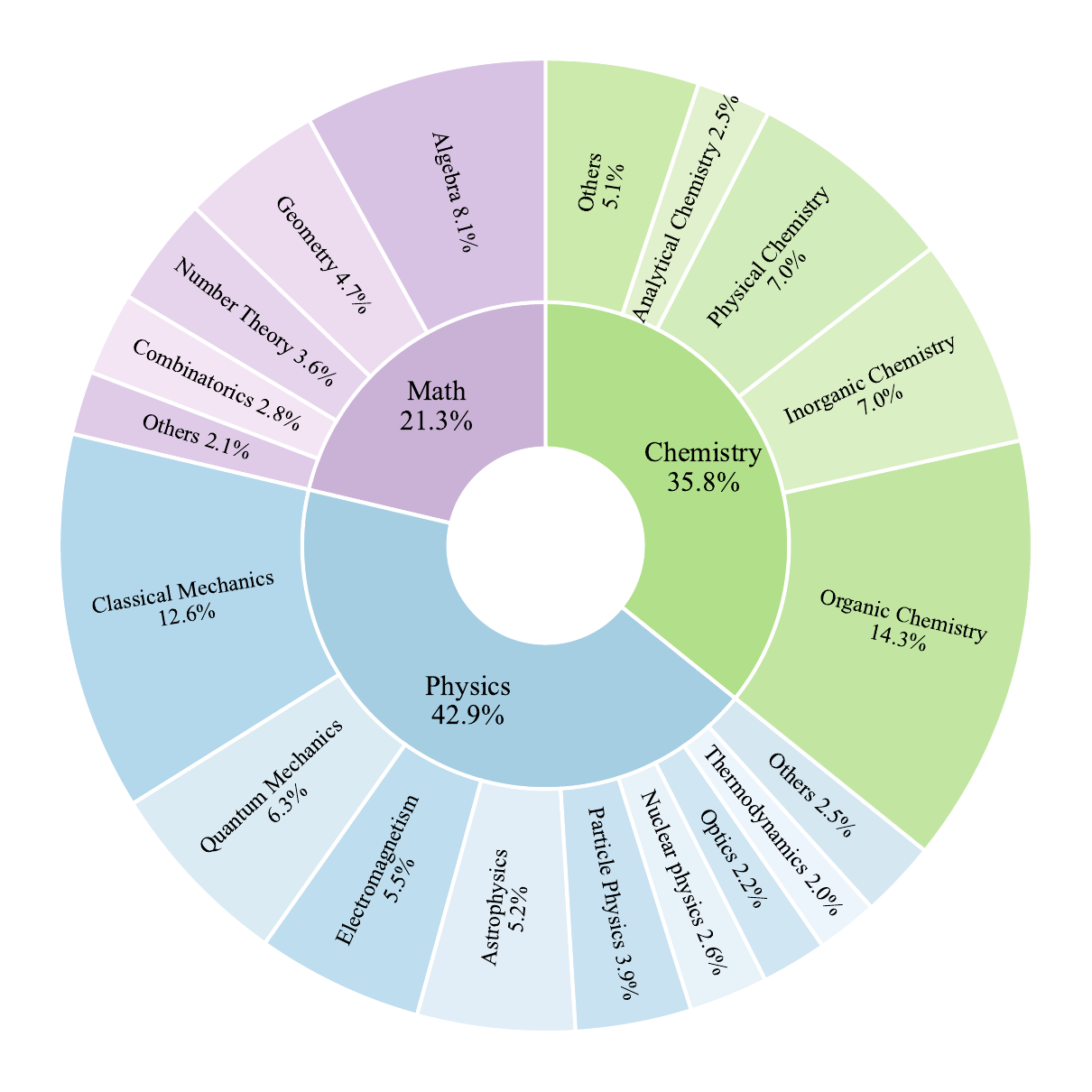}
        \caption{Category and Subcategory Distribution of Think-Bench.}
        \label{fig:distribution}
    \end{minipage}
    \hfill
    \begin{minipage}{0.4\textwidth}
        \centering 
        \begin{tabular}{lc}
            \toprule
            Statistic  & Number \\
            \midrule
            Total questions & 1375 \\
            \quad -Choice questions & 929 \\
            \quad -Free-form questions & 446 \\
            \quad -Math questions & 293 \\
            \quad -Physics questions & 590 \\
            \quad -Chemistry questions & 492 \\
            \midrule
            Total key step annotation & 13311 \\
            \quad -Average inference step & 9.68  \\
            \midrule
            Maximum question length & 1893 \\
            Maximum answer length & 372 \\
            Average question length & 422.42  \\
            Average answer length & 7.59  \\
            \bottomrule
        \end{tabular}
        \captionof{table}{Key Statistics of Think-Bench.}
        \label{tab:key_statistics}
    \end{minipage}
    \vspace{-20pt}
\end{figure*}

As shown in Figure~\ref{fig:datasetOverview}, Think-Bench is a dataset specifically designed to evaluate the thinking efficiency and the quality of CoTs of LRMs in complex reasoning tasks. This dataset comprises 1,375 carefully selected and organized data samples, covering three core subjects: mathematics, physics, and chemistry. Within each subject, the number of simple questions is approximately equal to the number of difficult questions. The data sources are diverse, drawing from multiple academic datasets, including MMLU~\citep{hendrycks2020measuring}, Math500~\citep{hendrycks2021measuring}, AGIEval~\citep{zhong2023agieval}, AIME~\citep{aime_1983_2024}, GPQA~\citep{rein2024gpqa}, SciKnowEval~\citep{feng2024sciknoweval}, and UGPhysics~\citep{xu2025ugphysics}.

\subsection{Data Collection}
During the construction of Think-Bench, we aggregated questions from multiple authoritative, publicly available datasets. The distribution of the Think-Bench across different disciplines is shown in Table~\ref{tab:key_statistics}. To ensure fairness, all samples were selected randomly. After the selection process, we conducted a systematic data cleaning and verification procedure to remove duplicate and invalid entries. The final dataset consists of 1,375 data points, after which we carried out the data annotation work. This benchmark covers the core disciplines of mathematics, physics, and chemistry, which inherently require structured and multi-step reasoning. Therefore, it provides a robust and rigorous foundation for evaluating the performance of reasoning models. Detailed statistics regarding the data composition can be found in Figure~\ref{fig:distribution} and Table~\ref{tab:key_statistics}.

\subsection{Data Annotation and Review}
To systematically evaluate the CoT reasoning capabilities of LRMs on reasoning tasks, we implement a fine-grained annotation framework for key reasoning steps across all questions. Key steps are defined as essential logical components that must be completed and cannot be omitted in the process of achieving the correct answer. The annotation process is implemented in the following steps: First, we leverage Claude 3.7 Sonnet to generate comprehensive reasoning chains for each question (see Prompt~\ref{box:key_step_prompt} in Appendix~\ref{prompt}), which serve as baseline references. Based on the generated reasoning paths, we identify and extract the critical steps involved.  Since a single question may admit multiple logically valid reasoning paths, all reasonable and logically consistent solutions are considered and included.
\begin{figure*}[!htbp]
    \centering
    \includegraphics[width=0.9\linewidth]{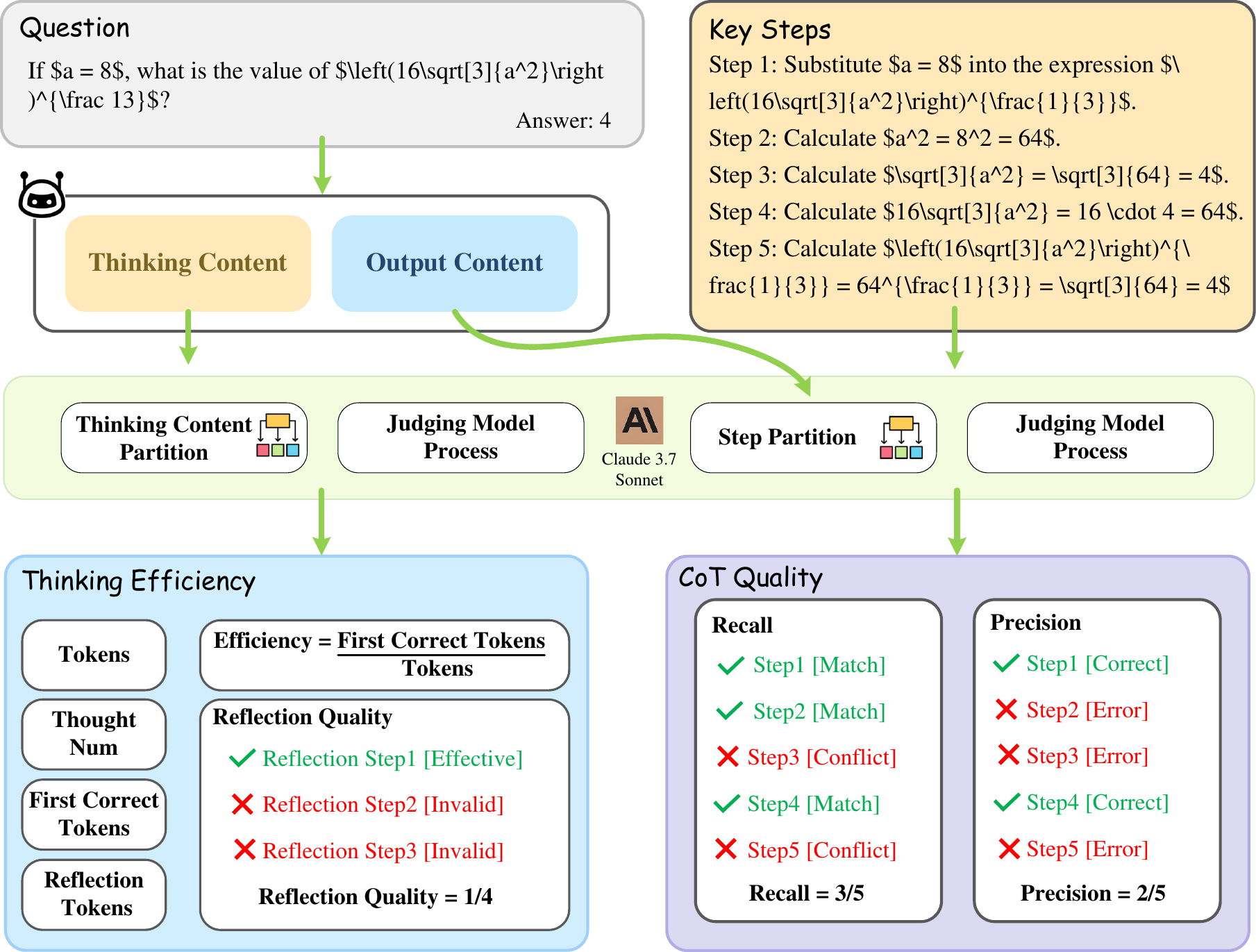}
    \caption{Illustration of Thinking Efficiency and CoT Quality Evaluation.}
    \label{fig:pipeline}
\end{figure*}

\begin{figure}[!ht]
    \centering
    \includegraphics[width=\linewidth]{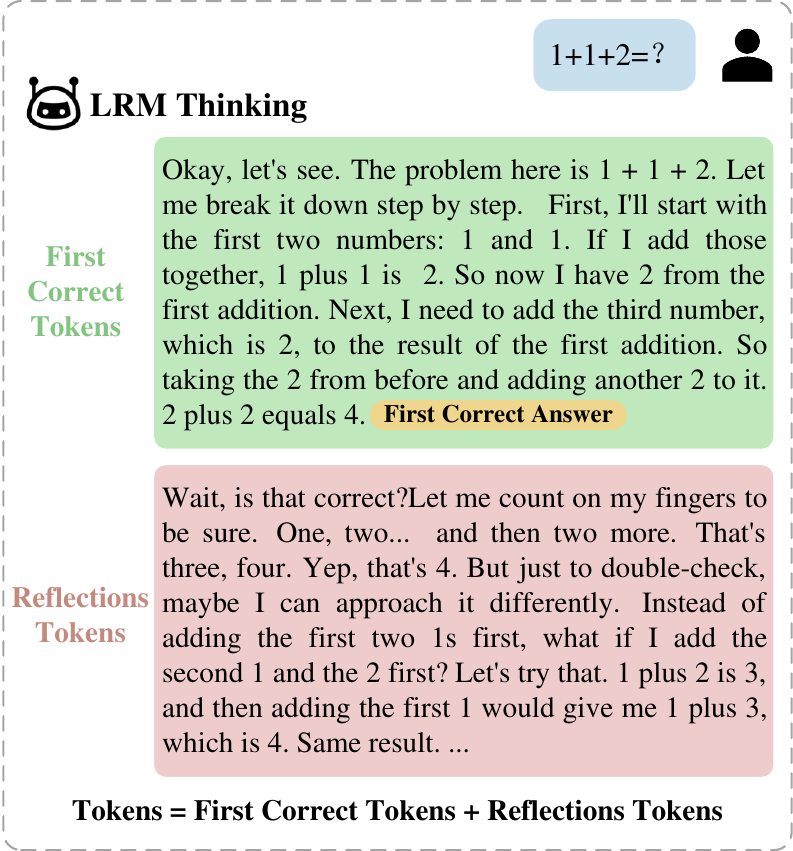}
    \caption{Example of Thinking Process Analysis in a LRM.}
    \label{fig:example_efficiency}
\end{figure}

\section{Evaluation Strategy}
Recent studies have revealed that LRMs frequently exhibit low reasoning efficiency~\citep{sui2025stop, chen2024not}.  However, there is currently a lack of systematic benchmarks to evaluate this issue comprehensively.  A detailed analysis of LRMs' reasoning processes is crucial for understanding their efficiency limitations and underlying challenges.  Furthermore, existing benchmarks primarily assess the final answers to reasoning questions, neglecting the intermediate CoT steps.  To bridge this gap, we propose a novel benchmark that jointly evaluates both the efficiency and quality of reasoning CoTs, thereby enabling a more holistic assessment of LRMs' reasoning capabilities.

\subsection{Thinking Efficiency Evaluation}
\label{efficiency}
With the rapid development of LRMs, their ability to handle complex multi-step reasoning has significantly advanced~\citep{xu2025towards}. Notable models like OpenAI-o1~\citep{zhong2024evaluation}, DeepSeek-R1~\citep{guo2025deepseek}, and Qwen3~\citep{yang2025qwen3} have attracted growing interest for their human-like capacity for extended, reflective reasoning. Through advanced long CoT and test-time scaling methods, these models iteratively evaluate multiple reasoning paths before finalizing answers~\citep{chen2025towards,muennighoff2025s1}.

However, as test-time scaling consumes increasing computational resources, a critical challenge has emerged in LRMs' inference behaviors: \textbf{Overthinking}. This refers to the model's persistent tendency to engage in excessive and repetitive reasoning, often producing reasoning chains that span hundreds of tokens even for simple tasks. While such elaborate verification is justified for complex problems, test-time scaling amplifies this behavior, causing unnecessary computational overhead and inefficiency during inference for simple inputs.

As illustrated in Figures~\ref{fig:pipeline} and~\ref{fig:example_efficiency}, we propose six complementary metrics to systematically assess reasoning efficiency across token usage, inference dynamics, and reflective quality.

\textbf{Tokens} measure the total token count processed before final prediction, representing reasoning chain length and providing a fundamental basis for computational cost estimation.

\textbf{First Correct Tokens} measures the token count from reasoning initiation until the first occurrence of a correct answer. This metric evaluates the model’s speed in reaching a valid solution during reasoning, where fewer tokens indicate faster correct convergence. The identification prompt is detailed in Prompt~\ref{box:correct_answer_prompt}.

\textbf{Efficiency} is a normalized metric that refers to the ratio of first correct tokens to the total number of reasoning tokens. Formally, it is defined as:

\begin{equation}
\text{Efficiency} = \frac{1}{N} \sum_{i=1}^{N} \frac{\hat{T}_i}{T_i}.
\end{equation}

where $\hat{T}_i$ denotes the number of tokens generated by the model before the first occurrence of the correct answer in its response, and let $T_i$ represent the total number of reasoning tokens for the $i$-th instance. If the model fails to produce a correct answer, we set $\hat{T}_i = 0$. A higher value of this metric indicates more efficient reasoning behavior. Concrete examples illustrating this calculation can be found in Figure~\ref{fig:efficiency_calculate} in Appendix~\ref{sec_efficiency_example}.

\textbf{Reflection Quality} measures the efficacy of the model's self-reflective reasoning, particularly after producing a correct answer. Not all reflective steps contribute meaningfully: some merely reiterate prior conclusions, while others may introduce erroneous revisions. We define a valid reflection as one that either (i) accurately identifies a prior error or (ii) provides new insights that confirm an earlier conclusion. Let \( R \) represent the total set of reflective steps, and \( R_{\text{valid}} \) denote the subset of valid reflections. The metric is defined as follows:

\begin{equation}
\text{Reflection Quality} = \frac{|R_{\text{valid}}|}{|R|}.
\end{equation}

This score quantifies the efficacy of the model's reflection process, where higher values indicate more meaningful self-verification behavior as opposed to producing redundant or counterproductive content. The prompt used to guide this reflection process is provided in Prompt~\ref{box:reflection_quality}.

\textbf{Reflection Tokens} quantify the token count generated from the first correct answer to the conclusion of the reasoning process. This segment typically encompasses verification steps, reflective analysis, and conclusion restatements. Although such content may provide valuable insights, excessive length often signals reasoning inefficiency or unnecessary repetition.

\textbf{Thought Num} measures how often the model changes reasoning paths. This metric is estimated by counting discourse markers like "alternatively," "on second thought," and "wait a moment." A higher count may indicate instability in reasoning or a tendency toward exploratory behavior.

\subsection{CoT Quality Evaluation}
\label{cot}
As LLMs increasingly adopt CoT reasoning strategies, assessing the quality of their internal reasoning processes has emerged as a critical research challenge~\citep{jiang2025mme}. Existing evaluation approaches predominantly focus on final answer accuracy~\citep{wang2019superglue, hendrycks2021measuring, suzgun2022challenging}, while largely overlooking the validity and robustness of intermediate reasoning steps. To bridge this gap, we adopt a reference-based evaluation framework, inspired by MME-CoT~\citep{jiang2025mme}. Our proposed framework measures the reasoning quality from two interpretable dimensions: \textbf{Recall} and \textbf{Precision}.

\begin{table*}[!htbp]
  \centering
    \resizebox{\linewidth}{!}{
    \begin{tabular}{lccccccccc}
    \toprule
     Model name & Efficiency & Recall & Precision & Accuracy & \makecell{Reflection\\Quality} & \makecell{Thought\\Num} & Tokens & \makecell{Useful\\Tokens} & \makecell{Reflection\\Tokens} \\
    \midrule
    Claude-3-7-sonnet & 49.61\% & 81.29\% & 86.26\% & 94.25\% & 76.49\% & \textbf{0.28}  & \textbf{942.82}  & \textbf{446.09}  & \textbf{496.73}  \\
    Deepseek-r1-distill-qwen-1.5b & 37.14\% & 47.10\% & 59.61\% & 62.91\% & 61.88\% & 8.00  & 3734.49  & 1268.36  & 2466.13  \\
    Deepseek-r1-distill-qwen-7b & 49.53\% & 63.65\% & 77.29\% & 68.51\% & 77.70\% & 9.42  & 3504.76  & 1641.91  & 1862.85  \\
    Deepseek-r1-distill-qwen-14b & 50.70\% & 61.04\% & 79.97\% & 70.18\% & 82.40\% & 7.04  & 2814.75  & 1413.09  & 1401.66  \\
    Deepseek-r1-distill-qwen-32b & 52.62\% & 64.17\% & 83.76\% & 75.93\% & 84.46\% & 6.27  & 2697.70  & 1352.93  & 1344.77  \\
    Deepseek-r1 & 48.96\% & 80.80\% & 88.33\% & 88.80\% & 90.92\% & 9.17  & 3795.19  & 1912.12  & 1883.07  \\
    Ernie-x1-turbo-32k & 47.02\% & 82.03\% & \textbf{88.67\%} & 89.89\% & 90.97\% & 12.75  & 4692.21  & 2221.32  & 2470.89  \\
    Grok-3-mini-beta & \textbf{61.69\%} & 81.56\% & 86.51\% & 91.85\% & 88.20\% & 0.38  & 1891.34  & 1169.05  & 722.29  \\
    Qwen3-235b-a22b & 46.14\% & \textbf{85.80\%} & 86.97\% & \textbf{94.91\%} & \textbf{92.16\%}
    & 13.35  & 4969.05  & 2448.29  & 2520.76  \\
    Qwq-plus & 44.58\% & 80.40\% & 85.08\% & 89.60\% & 89.67\% & 22.63  & 5738.37  & 2646.73  & 3091.64  \\
    Glm-z1-air & 47.41\% & 80.16\% & 83.18\% & 88.06\% & 89.17\% & 9.80  & 3678.68  & 1775.07  & 1903.61  \\
    \bottomrule
    \end{tabular}%
    }
  \caption{Evaluation of Nine Metrics on CoT and Efficiency in Think-Bench. Best performance in \textbf{bold}.}
  \label{tab:main_results}%
  \vspace{-1mm}
\end{table*}%

As illustrated in Figure~\ref{fig:pipeline}, each CoT response is decomposed into multiple reasoning steps through the prompt detailed in Prompt~\ref{box:precision_prompt}.
\[
R = \{r_1, r_2, \dots, r_M\}.
\]

To evaluate its quality, $R$ is compared against a pre-annotated reference set containing key reasoning components.
\[
S = \{s_1, s_2, \dots, s_N\}.
\]
Each $r_j$ is judged for semantic alignment with any $s_i$, using Claude 3.7 Sonnet as a judge guided by consistent prompting instructions. The prompt designed to extract matching steps for computing recall and precision is provided in Prompt~\ref{box:recall_prompt} and Prompt~\ref{box:precision_prompt} of Appendix~\ref{prompt}. We define:

\begin{itemize}
\item
$R_{\text{match}} \subseteq R$: the subset of reasoning steps in $R$ that correctly match at least one reference step in $S$.
\item 
$S_{\text{covered}} \subseteq S$: the subset of reference steps that are successfully matched by at least one step in $R$.
\end{itemize}

Using $R_{\text{match}}$ and $S_{\text{covered}}$, we compute the \textbf{Recall} and \textbf{Precision} metrics as follows:

\begin{equation}
\textbf{Recall} = \frac{|S_{\text{covered}}|}{|S|}
\end{equation}

\begin{equation}
\textbf{Precision} = \frac{|R_{\text{match}}|}{|R|}
\end{equation}

Recall measures the extent to which essential reasoning steps are accurately captured in the LRM's output, reflecting the informativeness and comprehensiveness of the generated reasoning chain. In contrast, precision evaluates the correctness and relevance of the reasoning steps, penalizing any instance of inaccuracy or logical inconsistency.
\section{Experiment}
\subsection{Experimental Setup}

\paragraph{Evaluation Models}
To systematically evaluate both the efficiency and quality of reasoning with CoT in LRMs, we select eleven representative models spanning diverse architectures and parameter scales. Our evaluation encompasses both proprietary and open-source LRMs. Specifically, we include Claude 3.7 Sonnet~\citep{anthropic2025claude37}, a proprietary model widely recognized for its strong performance in multi-turn reasoning tasks. We also conduct a comprehensive assessment of the DeepSeek-R1 family~\citep{guo2025deepseek}, including the full-scale DeepSeek-R1 and its distilled Qwen-1.5-based variants at 1.5B, 7B, 14B, and 32B scales, all explicitly optimized for efficient multi-step reasoning. Additionally, we evaluate Qwen3-235B-A22B~\citep{qwen3} and Qwq-Plus~\citep{QwQ2024Nov}, both equipped with reflection and alignment mechanisms to support long-context inference. To further explore model behavior under extended reasoning conditions, we include Ernie-X1-Turbo-32K~\citep{erniex1}, optimized for long input sequences, along with Grok-3-Mini-Beta~\citep{xai2025grok3} and GLM-Z1-Air~\citep{glm2024chatglm}.

\paragraph{Implementation Details}
Throughout the evaluation process, we initially employed the tested LRMs to generate responses to the entries from Think-Bench.  All other model hyperparameters followed default settings unless otherwise specified. Subsequently, Claude 3.7 Sonnet was utilized to analyze the reasoning steps and underlying thinking processes of these responses. The detailed prompt used for the analysis with Claude 3.7 Sonnet is provided in Appendix~\ref{prompt}. Finally, we computed our proposed evaluation metrics to assess the thinking efficiency and reliability of the tested LRMs.

\subsection{Quantitative Results}
We conduct a comprehensive evaluation of LRMs using our proposed Think-Bench. The main results are presented in Tables~\ref{tab:main_results} and Tables~\ref{tab:question_type}. We begin with an analysis of the overall performance, followed by an in-depth discussion of the key findings.

\begin{table*}[!htbp]
  \centering
  \renewcommand\tabcolsep{3pt} 
  \renewcommand\arraystretch{1.2} 
  \resizebox{\linewidth}{!}{ 
  \begin{tabular}{l|cc|cc|cc|cc|cc}
    \toprule
    \multirow{2}[2]{*}{Model name} & \multicolumn{2}{c|}{Recall} & \multicolumn{2}{c|}{Precision} & \multicolumn{2}{c|}{Reflection Quality} & \multicolumn{2}{c|}{Tokens} & \multicolumn{2}{c}{Efficiency} \\
          & Simple   & Difficult  & Simple   & Difficult  & Simple   & Difficult  & Simple   & Difficult  & Simple   & Difficult \\
    \midrule
    Claude-3-7-sonnet & 88.49\% & 74.05\% & 92.94\% & 79.56\% & 92.95\% & 90.29\% & \textbf{673.24}  & \textbf{1216.01}  & 0.52  & 0.47  \\
    Deepseek-r1-distill-qwen-1.5b & 55.74\% & 38.42\% & 69.35\% & 49.84\% & 73.22\% & 68.55\% & 2149.94  & 5325.97  & 0.40  & 0.34  \\
    Deepseek-r1-distill-qwen-7b & 70.75\% & 51.30\% & 88.84\% & 71.06\% & 90.09\% & 83.64\% & 1575.27  & 4059.64  & 0.51  & 0.51  \\
    Deepseek-r1-distill-qwen-14b & 72.99\% & 55.31\% & 92.55\% & 74.93\% & 91.43\% & 85.07\% & 1514.53  & 3886.05  & 0.52  & 0.53  \\
    Deepseek-r1-distill-qwen-32b & 69.88\% & 57.39\% & 85.79\% & 68.75\% & 84.99\% & 78.91\% & 2074.32  & 4941.47  & 0.49  & 0.50  \\
    Deepseek-r1 & 88.80\% & 72.77\% & \textbf{95.54\%} & 81.09\% & 95.15\% & 90.00\% & 2058.35  & 5539.63  & 0.46  & 0.52  \\
    Ernie-x1-turbo-32k & 90.26\% & 73.76\% & 95.44\% & \textbf{81.88\%} & 95.00\% & 89.09\% & 2679.32  & 6713.91  & 0.43  & 0.51  \\
    Grok-3-mini-beta & 89.09\% & 74.01\% & 93.47\% & 79.51\% & 93.17\% & 88.78\% & 1242.27  & 2543.25  & \textbf{0.60}  & \textbf{0.63}  \\
    Qwen3-235b-a22b & \textbf{92.87\%} & \textbf{78.70\%} & 95.33\% & 78.57\% & \textbf{96.29\%} & \textbf{91.43\%} & 2818.69  & 7128.80  & 0.42  & 0.50  \\
    Qwq-plus & 90.04\% & 70.72\% & 94.89\% & 75.23\% & 94.60\% & 87.42\% & 3289.45  & 8197.99  & 0.41  & 0.48  \\
    GLM-Z1-Air & 88.69\% & 71.60\% & 93.31\% & 72.99\% & 94.79\% & 86.55\% & 1931.36  & 5433.63  & 0.45  & 0.49  \\
    \bottomrule
  \end{tabular}
  }
  \caption{Evaluation Results of CoT and Efficiency in Think-Bench Classified by Difficulty Levels. Best performance in \textbf{bold}.}
  \label{tab:question_type}
  \vspace{-5pt}
\end{table*}

\paragraph{Overall Performance}
To comprehensively evaluate the performance of LRMs, we report results across two key dimensions: efficiency and CoT quality, including our proposed efficiency metrics, recall, precision and accuracy, as shown in Table~\ref{tab:main_results}.  Our analysis shows that while there exists a consistent trade-off between token usage and reasoning performance, different models exhibit significant variability in their inference behaviors.

In terms of efficiency, Grok-3-mini-beta achieves the highest score of 61.69\%, followed by Deepseek-r1-distill-qwen-32b at 52.62\% and Deepseek-r1-distill-qwen-14b at 50.70\%, indicating a more economical use of tokens to reach correct answers.  In contrast, larger models such as Qwen3-235b-a22b and Qwq-plus exhibit lower efficiency, scoring 46.14\% and 44.58\% respectively.  This decrease in efficiency is attributed to their prolonged reasoning chains, despite having strong CoT quality.

Regarding CoT quality, Qwen3-235b-a22b and Ernie-x1-turbo-32k stand out by achieving the highest reflection quality scores, with values of 92.16\% and 90.97\%, respectively. They also demonstrate top-tier precision at 86.97\% and 88.67\%, and recall rates of 85.80\% and 82.03\%. These impressive results highlight the advantages of large-scale models with reflection-enhanced reasoning capabilities, which not only lead to accurate conclusions but also enable reliable verification and correction processes. In contrast, smaller distilled models, such as Deepseek-r1-distill-qwen-1.5b, perform poorly across all quality metrics, particularly in precision (59.61\%) and recall (47.10\%).

An important behavioral indicator is Thought Num, reflecting how often the model switches or reconsiders its CoT. Qwq-plus shows the highest value (22.63), indicating frequent reflective iterations. However, such reflections don’t always lead to better performance and may reduce efficiency. In contrast, models like Claude 3.7 Sonnet and Grok-3-mini-beta maintain very low Thought Num values (0.28 and 0.38) while still achieving a balanced and high-quality reasoning process.

Regarding token consumption, Qwq-plus and Qwen3-235b-a22b each use over 4,900 tokens per response, with a substantial portion from reflection (3,091.64 and 2,520.76, respectively). This suggests tendencies toward overthinking. In contrast, Claude-3-7-Sonnet completes its reasoning in under 1,000 tokens, demonstrating concise and effective inference with minimal redundancy.

\begin{figure*}
    \centering
    \includegraphics[width=\linewidth]{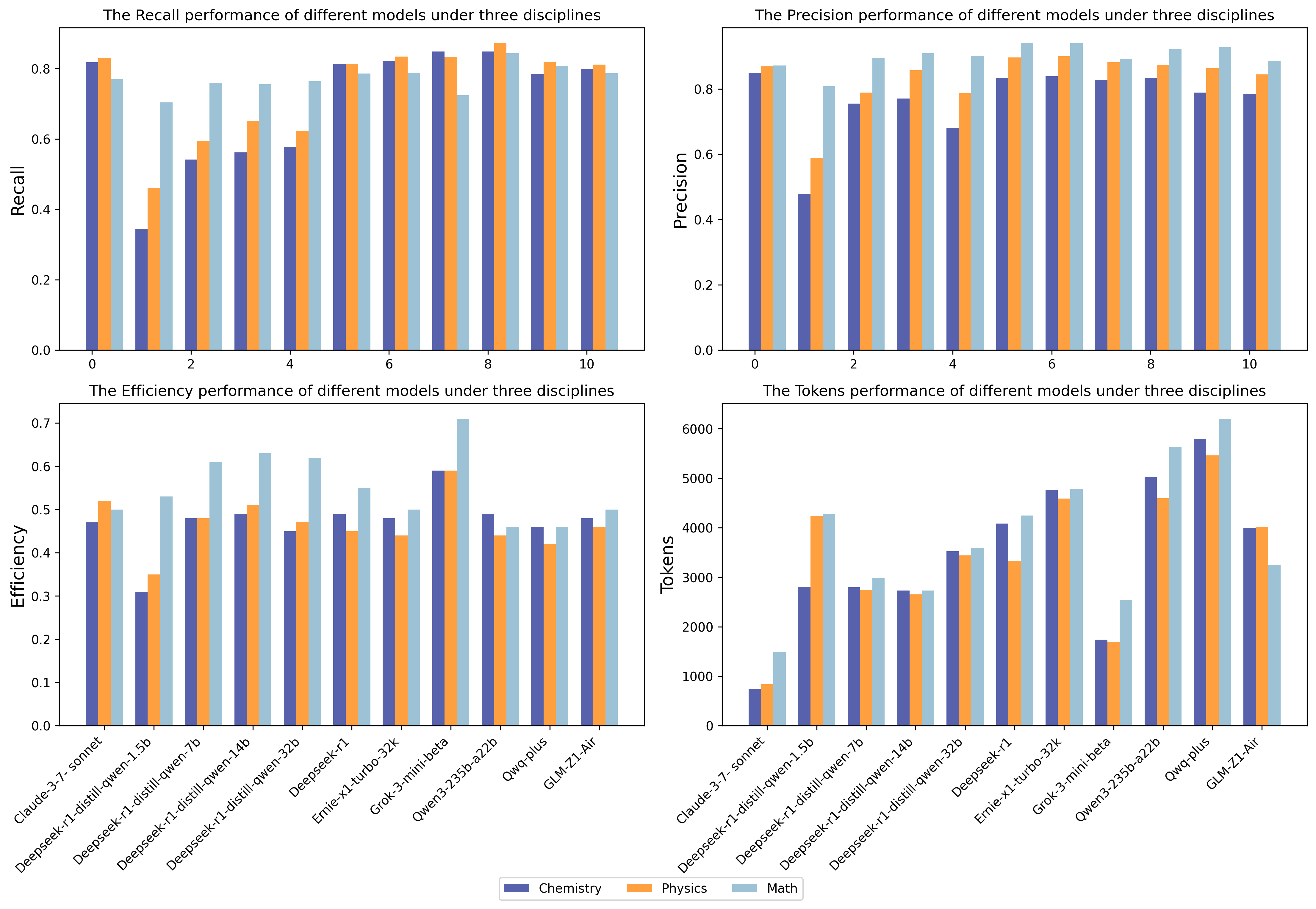}
    \caption{Comparative Performance of Models in Chemistry, Physics, and Math.}
    \label{fig:different_category}
    \vspace{-1mm}
\end{figure*}

\paragraph{Evaluation and Analysis Based on Difficulty Levels of Questions}

The results in Table~\ref{tab:question_type} show that most LRMs demonstrate significantly lower average efficiency on simple questions than on difficult ones. This suggests that when faced with simple questions, these models tend to overthink and generate unnecessary reasoning chains. In contrast, for high difficulty questions, the models focus more effectively, eliminating redundant reasoning steps and improving efficiency. Additionally, token consumption for difficult questions is consistently higher than for simple ones, due to the extra inferential steps needed to tackle complex problems. As reflection quality, recall, and precision all decline slightly as task difficulty increases, this illustrates that while difficult questions require more computational resources, they present greater challenges to the model's reasoning capabilities.

\paragraph{Evaluation and Analysis Based on Different Subjects}

As shown in Figure~\ref{fig:different_category}, the multi-disciplinary evaluation in Think-Bench reveals notable performance differences across chemistry, physics, and mathematics. Mathematical tasks generally lead to higher token consumption and lower reasoning efficiency, even for strong models, suggesting a reliance on lengthy CoTs and structured outputs. In contrast, chemistry and physics tasks typically exhibit better efficiency and lower token usage.

Regarding CoT quality evaluation, the recall and precision generally show a positive correlation in various disciplinary tasks within Think-Bench, but there are also obvious structural differences. Specifically, in chemistry and physics tasks, the precision rate of the model is often significantly better than the recall rate.  This phenomenon reflects that the current LRMs' strategy in generating answers for reasoning questions is relatively conservative, such that it is more inclined to output answers with high confidence.

\subsection{Error Analysis}
During the evaluation experiment, we observed that some models, particularly the distillation models from the DeepSeek series and the ERNIE-X1-Turbo-32K, exhibited an unusual issue of generating empty outputs. This issue primarily manifested in the model generating only intermediate reasoning content without producing a final answer for certain questions. See Figure~\ref{fig:error_example} in Appendix~\ref{error_example} for a concrete example. Potential causes include limitations in their inference mechanisms, context processing capabilities, or deployment implementations. As the issue was sporadic and difficult to reproduce systematically, we chose to automatically skip any samples with invalid outputs to preserve the integrity of the evaluation and ensure the stability of subsequent analyses.
\section{Conclusion}
This paper presents the Think-Bench dataset, a benchmark designed to systematically evaluate the reasoning efficiency and CoT quality of LRMs. The dataset consists of tasks from three disciplines: mathematics, physics, and chemistry. Each task is provided at two difficulty levels: Simple and Difficult.    Evaluation is conducted using nine metrics, including six efficiency indicators, two CoT quality measures and accuracy. To verify the effectiveness of Think-Bench and to assess the reasoning efficiency and CoT quality of mainstream LRMs, we conduct a comprehensive evaluation of 11 representative models.    Experimental results show that most models exhibit overthinking behaviors on simple questions, generating excessive reasoning tokens and leading to unnecessary computational overhead.
This study not only highlights the limitations of current LRMs in their use of computational resources in inference time, but also offers insights that may inspire future research, including designing dynamic reasoning pathways, early exit mechanisms, and enhancing adaptability across disciplines.

\section*{Limitations}
The Think-Bench benchmark proposed in this study currently covers only three disciplines: mathematics, physics, and chemistry, which limits its effectiveness in evaluating models’ reasoning abilities across a broader range of subjects or in more complex real-world scenarios. Furthermore, the evaluation process relies on large language models to assess the reasoning steps of target models, introducing potential variability due to the performance and stability of the judging models themselves. In this study, Claude 3.7 Sonnet is utilized both as the model being evaluated and as the judging model, which may further enhance this issue by introducing bias in the assessment. In addition, although the dataset annotation attempts to incorporate a variety of valid reasoning paths, it is challenging to exhaust all possible solution strategies, which may lead to incomplete evaluations of models that adopt reasonable but unannotated reasoning approaches.


\bibliography{acl_latex}

\appendix

\section{Related Work}
\label{appendix:related_work}
In recent years, evaluating the reasoning capabilities of LLMs has become a pivotal research focus within the field of natural language processing\citep{chang2024survey}. Existing evaluation methodologies can be broadly categorized into two approaches: outcome-oriented and process-oriented assessments.

Outcome-oriented evaluations primarily emphasize the accuracy of the model's final output. Prominent benchmarks in this category include SuperGLUE \citep{wang2019superglue}, MMLU \citep{hendrycks2021measuring}, and BIG-bench \citep{suzgun2022challenging}. These benchmarks encompass a wide array of tasks, ranging from language comprehension to domain-specific question answering, thereby standardizing the performance assessment of LLMs. However, such methods often overlook the interpretability and rationality of the model's reasoning process, particularly in complex problem-solving scenarios where the significance of intermediate steps is substantially undervalued.

To address these limitations, process-oriented evaluation methodologies have been introduced\citep{zheng2024processbench, jiang2025mme}. The CoT reasoning framework \citep{wei2022chain} exemplifies this approach by explicitly guiding models to generate intermediate reasoning steps, thereby enhancing performance in mathematical and logical tasks. Subsequent studies, such as Auto-CoT \citep{zhang2022automatic}, Tree-of-Thought \citep{yao2023tree}, and ReAct \citep{yao2023react}, have further augmented the flexibility and diversity of reasoning pathways. 

Furthermore, the evaluation of multidisciplinary reasoning capabilities has become a focal point in current research. Researchers have developed various assessment benchmarks and methodologies to comprehensively measure the reasoning abilities of LLMs across different academic disciplines. For instance, the Advanced Reasoning Benchmark (ARB) is a comprehensive reasoning benchmark that spans multiple domains, including mathematics, physics, biology, chemistry, and law, designed to evaluate the performance of LLMs in complex reasoning tasks \citep{sawada2023arb}. Multi-LogiEval is a dataset that provides an integrated evaluation of LLMs' multi-step logical reasoning abilities, covering multiple types of logic such as propositional logic, first-order logic, and non-monotonic logic \citep{patel2024multi}.

Additionally, large reasoning models tend to exhibit overthinking behavior during chain-of-thought reasoning, where excessively long and unnecessary reasoning steps are generated even for simple or ill-posed problems \citep{sui2025stop, chen2024not, fan2025missing, pu2025thoughtterminator}. This phenomenon is often attributed to the models' lack of proper termination mechanisms and insufficient confidence estimation, leading to inefficient inference and degraded accuracy. It has been observed that the issue becomes more pronounced when essential premises are missing from the input \citep{fan2025missing}. To address this, several approaches have been proposed, including the introduction of new reasoning efficiency metrics and self-training strategies that encourage concise reasoning \citep{chen2024not}, as well as dynamic early-exit mechanisms that halt inference when sufficient confidence is reached \citep{yang2025dynamic}. Additionally, path scoring methods have been developed to prefer less redundant reasoning paths, thereby improving performance while reducing computational cost \citep{cuadron2025danger}.

\section{More experimental results}
\label{more_results}
The comparative analysis of Tables~\ref{tab:category1} and Tables~\ref{tab:category2} highlights the trade-offs between reasoning quality and efficiency across various disciplines. In mathematics, top-performing models, such as Qwen3-235b-a22b and Qwq-plus, achieve high quality of reflection and precision but require a large number of tokens. In contrast, Grok-3-mini-beta strikes a balance between conciseness and accuracy, achieving a precision of 88.2\% in physics and a recall of 84.9\% in chemistry while using fewer tokens. Furthermore, smaller distilled variants, like Deepseek-r1-distill-qwen-1.5b, demonstrate significant limitations in domain-specific reasoning, particularly in physics and chemistry, where both recall and precision fall below 50\%.

\begin{table*}[!htbp]
  \centering
  \vspace{-30pt}
  \renewcommand\tabcolsep{3pt} 
  \renewcommand\arraystretch{1.2} 
  \resizebox{\linewidth}{!}{ 
    \begin{tabular}{l|ccc|ccc|ccc|ccc}
    \toprule
    \multicolumn{1}{l|}{\multirow{2}[2]{*}{Model name}} & \multicolumn{3}{c|}{Recall} & \multicolumn{3}{c|}{Precision} & \multicolumn{3}{c|}{Reflection Quality} & \multicolumn{3}{c}{Tokens} \\
          & Chemistry & Physics & Math  & Chemistry & Physics & Math  & Chemistry & Physics & Math  & Chemistry & Physics & Math \\
    \midrule
    C-3-7-sonnet & 81.83\% & 82.97\% & 76.97\% & 84.92\% & 86.94\% & 87.16\% & 87.89\% & 90.39\% & 95.25\% & 744.21 & 836.63 & 1494.49 \\
    Ds-r1-distill-qwen-1.5b & 34.43\% & 46.11\% & 70.37\% & 47.91\% & 58.86\% & 80.78\% & 59.37\% & 70.18\% & 89.20\% & 2807.87 & 4236.72 & 4279.11 \\
    Ds-r1-distill-qwen-7b & 54.11\% & 59.42\% & 75.95\% & 75.55\% & 78.93\% & 89.49\% & 80.16\% & 87.28\% & 95.58\% & 2800.44 & 2742.55 & 2984.15 \\
    Ds-r1-distill-qwen-14b & 56.21\% & 65.16\% & 75.54\% & 77.07\% & 85.76\% & 90.95\% & 80.56\% & 90.89\% & 96.36\% & 2729.47 & 2654.41 & 2731.53 \\
    Ds-r1-distill-qwen-32b & 57.75\% & 62.24\% & 76.39\% & 68.02\% & 78.67\% & 90.09\% & 72.12\% & 83.86\% & 94.46\% & 3525.06 & 3442.9 & 3595.26 \\
    Ds-r1 & 81.39\% & 81.41\% & 78.59\% & 83.34\% & 89.63\% & 94.09\% & 87.79\% & 93.32\% & 98.05\% & 4082.93 & 3331.74 & 4245.25 \\
    Es-x1-turbo-32k & 82.25\% & 83.44\% & 78.82\% & 83.92\% & 90.00\% & 94.00\% & 87.95\% & 92.75\% & 97.00\% & 4762.48 & 4588.33 & 4783.4 \\
    G-3-mini-beta & 84.86\% & 83.34\% & 72.45\% & 82.79\% & 88.22\% & 89.30\% & 86.04\% & 91.69\% & 96.53\% & 1742.1 & 1690.57 & 2546.22 \\
    Q3-235b-a22b & 84.88\% & 87.27\% & 84.37\% & 83.39\% & 87.36\% & 92.18\% & 90.29\% & 94.55\% & 98.05\% & 5022.55 & 4592.91 & 5636.6 \\
    Qwq-plus & 78.38\% & 81.92\% & 80.73\% & 78.87\% & 86.42\% & 92.80\% & 84.76\% & 92.29\% & 98.11\% & 5795.2 & 5460.61 & 6202.25 \\
    Glm-z1-air & 79.93\% & 81.09\% & 78.66\% & 78.37\% & 84.49\% & 88.66\% & 84.51\% & 91.84\% & 96.97\% & 3992.9 & 4013.77 & 3250.19 \\
    \bottomrule
    \end{tabular}
    }
  \caption{\textbf{Comparative Performance of Models in Different Category.} C-3-7-sonnet: claude 3.7 sonnet; Ds-r1-distill-qwen-1.5b: deepseek-r1-distill-qwen-1.5b; Ds-r1-distill-qwen-7b: deepseek-r1-distill-qwen-14b; Ds-r1-distill-qwen-14b: deepseek-r1-distill-qwen-32b; Ds-r1-distill-qwen-32b: deepseek-r1-distill-qwen-7b; Ds-r1: deepseek-reasoner; Es-x1-turbo-32k: ernie-x1-turbo-32k; G-3-mini-beta: grok-3-mini-beta; Q3-235b-a22b: qwen3-235b-a22b.}
  \label{tab:category1}%
\end{table*}%

\begin{table*}[!htbp]
  \centering
  \renewcommand\tabcolsep{3pt} 
  \renewcommand\arraystretch{1.2} 
  \resizebox{\linewidth}{!}{ 
    \begin{tabular}{l|ccc|ccc|ccc|ccc}
    \toprule
    \multicolumn{1}{l|}{\multirow{2}[2]{*}{Model name}} & \multicolumn{3}{c|}{Thought Num} & \multicolumn{3}{c|}{Efficiency} & \multicolumn{3}{c|}{Useful Tokens} & \multicolumn{3}{c}{Reflection Tokens} \\
          & Chemistry & Physics & Math  & Chemistry & Physics & Math  & Chemistry & Physics & Math  & Chemistry & Physics & Math \\
    \midrule
    C-3-7-sonnet & 0.16  & 0.23  & 0.58  & 0.47  & 0.52  & 0.5   & 339.85 & 411.6 & 695.93 & 404.35 & 425.03 & 798.55 \\
    Ds-r1-distill-qwen-1.5b & 8.2   & 8.65  & 6.35  & 0.31  & 0.35  & 0.53  & 842.08 & 1243.98 & 2033.23 & 1965.79 & 2992.74 & 2245.88 \\
    Ds-r1-distill-qwen-7b & 9.76  & 6.13  & 4.3   & 0.48  & 0.48  & 0.61  & 1367.99 & 1243.05 & 1831.2 & 1432.45 & 1499.5 & 1152.94 \\
    Ds-r1-distill-qwen-14b & 8.24  & 5.75  & 4     & 0.49  & 0.51  & 0.63  & 1270.29 & 1226.93 & 1745.42 & 1459.18 & 1427.48 & 986.11 \\
    Ds-r1-distill-qwen-32b & 12.49 & 8.43  & 6.25  & 0.45  & 0.47  & 0.62  & 1535.14 & 1480.06 & 2147.1 & 1989.91 & 1962.84 & 1448.16 \\
    Ds-r1 & 13.13 & 6.96  & 6.97  & 0.49  & 0.45  & 0.55  & 2087.08 & 1521.81 & 2404.29 & 1995.86 & 1809.93 & 1840.97 \\
    Es-x1-turbo-32k & 17.14 & 11.15 & 8.62  & 0.48  & 0.44  & 0.5   & 2315.51 & 2031.88 & 2444.63 & 2446.98 & 2556.44 & 2338.77 \\
    G-3-mini-beta & 0.45  & 0.24  & 0.52  & 0.59  & 0.59  & 0.71  & 1049.76 & 985.93 & 1738.09 & 692.34 & 704.65 & 808.13 \\
    Q3-235b-a22b & 16.95 & 11.21 & 11.62 & 0.49  & 0.44  & 0.46  & 2547.58 & 2087.76 & 3007.54 & 2474.97 & 2505.15 & 2629.06 \\
    Qwq-plus & 30.98 & 18.24 & 17.44 & 0.46  & 0.42  & 0.46  & 2664.93 & 2351.91 & 3209.83 & 3130.26 & 3108.69 & 2992.42 \\
    Glm-z1-air & 14.11 & 6.99  & 7.61  & 0.48  & 0.46  & 0.50  & 1887.16 & 1483.43 & 2174.12 & 2105.78 & 1766.76 & 1839.66 \\
    \bottomrule
    \end{tabular}
    }
  \caption{\textbf{Comparative Performance of Models in Different Category.} C-3-7-sonnet: claude 3.7 sonnet; Ds-r1-distill-qwen-1.5b: deepseek-r1-distill-qwen-1.5b; Ds-r1-distill-qwen-7b: deepseek-r1-distill-qwen-14b; Ds-r1-distill-qwen-14b: deepseek-r1-distill-qwen-32b; Ds-r1-distill-qwen-32b: deepseek-r1-distill-qwen-7b; Ds-r1: deepseek-reasoner; Es-x1-turbo-32k: ernie-x1-turbo-32k; G-3-mini-beta: grok-3-mini-beta; Q3-235b-a22b: qwen3-235b-a22b.}
  \label{tab:category2}%
\end{table*}%

\begin{table*}[!htbp]
  \centering
  \renewcommand\tabcolsep{3pt} 
  \renewcommand\arraystretch{1.2} 
  \resizebox{0.8\linewidth}{!}{ 
    \begin{tabular}{l|cc|cc|cc}
    \toprule
    \multicolumn{1}{l|}{\multirow{2}[2]{*}{Model name}} & \multicolumn{2}{c|}{Thought Num} & \multicolumn{2}{c|}{Useful Tokens} & \multicolumn{2}{c}{Reflection Tokens} \\
          & Simple   & Difficult  & Simple   & Difficult  & Simple   & Difficult \\
    \midrule
    Claude-3-7-sonnet & 0.17  & 0.39  & 340.05 & 553.55 & 333.19 & 662.46 \\
    Deepseek-r1-distill-qwen-1.5b & 4.10  & 11.91 & 818.28 & 1720.40 & 1331.65 & 3605.56 \\
    Deepseek-r1-distill-qwen-7b & 3.05  & 11.04 & 825.64 & 2003.10 & 749.63 & 2056.54 \\
    Deepseek-r1-distill-qwen-14b & 2.86  & 9.70  & 777.32 & 1931.06 & 737.21 & 1954.99 \\
    Deepseek-r1-distill-qwen-32b & 4.68  & 14.18 & 960.86 & 2325.93 & 1113.45 & 2615.53 \\
    Deepseek-r1 & 4.26  & 14.11 & 950.72 & 2877.72 & 1107.63 & 2661.91 \\
    Ernie-x1-turbo-32k & 6.75  & 18.78 & 1144.62 & 3302.73 & 1534.70 & 3411.17 \\
    Grok-3-mini-beta & 0.22  & 0.54  & 765.60 & 1574.26 & 476.67 & 968.99 \\
    Qwen3-235b-a22b & 7.93  & 18.80 & 1245.45 & 3656.38 & 1573.24 & 3472.42 \\
    Qwq-plus & 11.09 & 34.21 & 1328.11 & 3971.12 & 1961.34 & 4226.88 \\
    Glm-z1-air & 4.27  & 15.36  & 876.64  & 2677.43  & 1054.72  & 2756.20  \\
    \bottomrule
    \end{tabular}
    }
  \caption{\textbf{Effect of question difficulty on other efficiency measures.}}
  \label{tab:appendix_ques_type}%
\end{table*}%

As shown in Table~\ref{tab:appendix_ques_type}. On simple tasks, Grok-3-mini-beta demonstrates efficient and focused reasoning, producing only 0.54 thoughts and consuming 1,574.26 tokens.  In contrast, when tackling difficult questions, larger models such as Qwen3-235b-a22b and Ernie-x1-turbo-32k generate over 3,600 tokens on average while achieving high reflection quality, reaching 92.16\% and 94.00\% respectively.  However, this increase in quality comes with reduced efficiency.  For example, Qwq-plus achieves only 44.58\% efficiency due to its high reflection token count after the answer, totalling 4,226.88 tokens.

\clearpage
\onecolumn
\section{More Qualitative Examples}
\label{appendix:example}
\subsection{Example of Computational Efficiency}
\label{sec_efficiency_example}
\begin{figure}[H]
    \centering
    \includegraphics[width=0.9\linewidth]{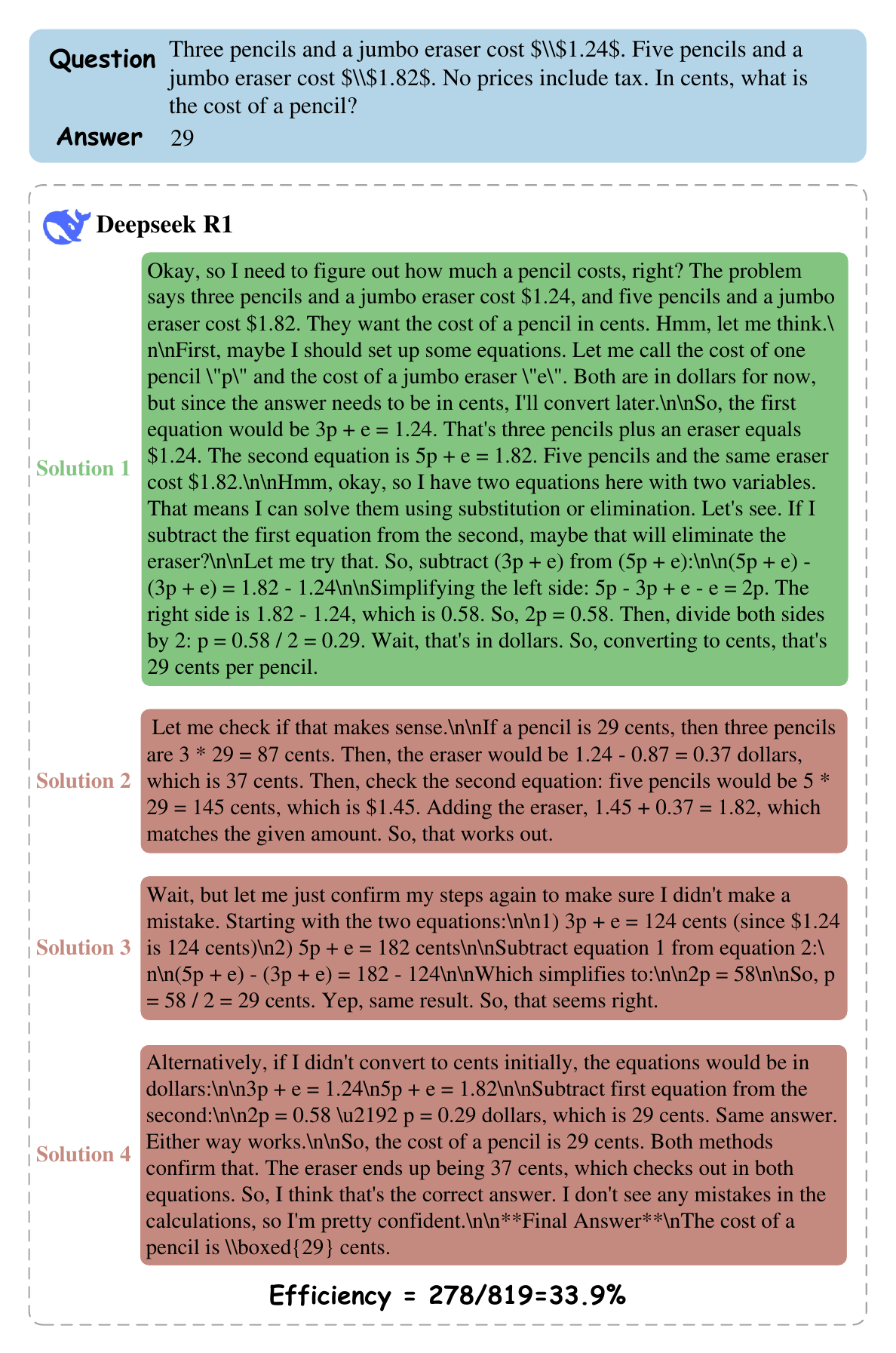}
    \caption{\textbf{Examples of Efficiency Evaluation.}}
    \label{fig:efficiency_calculate}
\end{figure}

In Figure~\ref{fig:efficiency_calculate}, we illustrate an example of efficient reasoning by an LRM.

\subsection{Error Example}
\label{error_example}
\begin{figure}[H]
    \centering
    \includegraphics[width=0.8\linewidth]{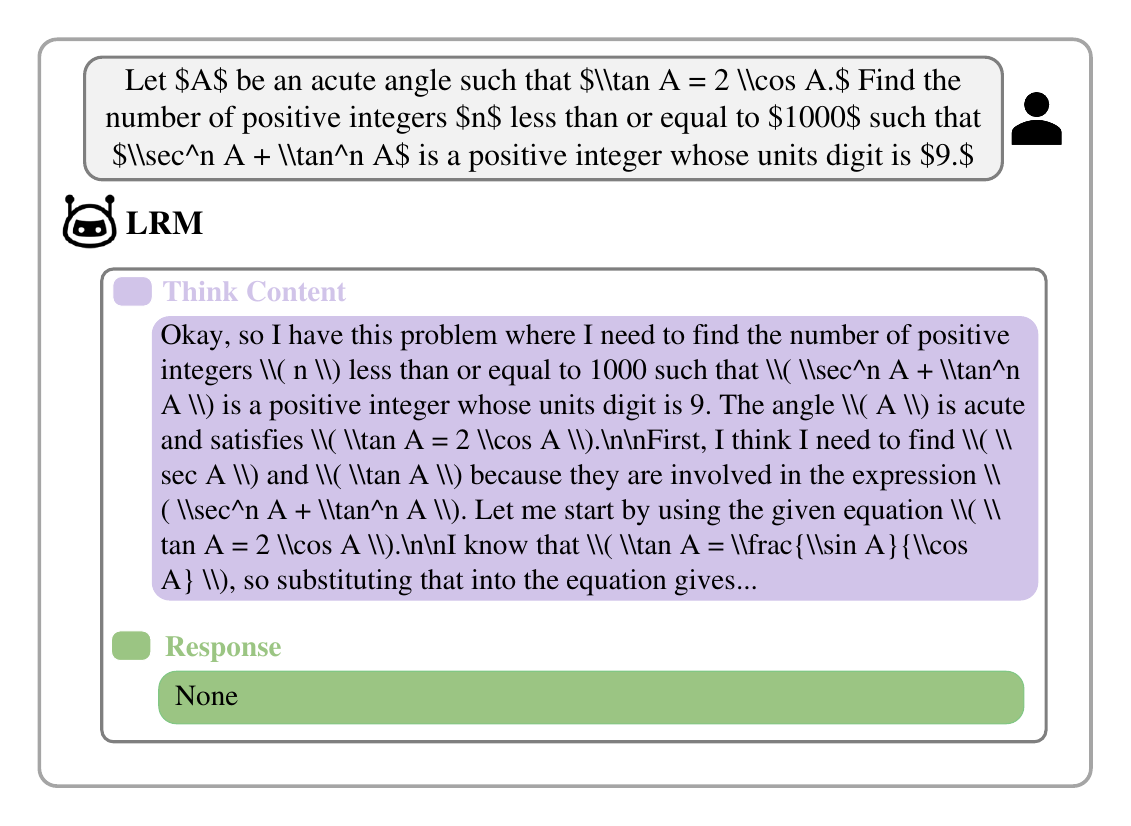}
    \caption{Failure Case Example in the Deepseek-r1-distill-qwen-32b Answering Process.}
    \label{fig:error_example}
\end{figure}

We present an example of an LRM output error in Figure~\ref{fig:error_example}.

\section{Evaluation Prompts}
\label{prompt}
\refstepcounter{promptbox}
\begin{tcolorbox}[breakable, colback=gray!5!white, colframe=gray!75!black, 
title=Key Steps Extraction Prompt, boxrule=0.5mm, width=\textwidth, arc=3mm, auto outer arc]
\label{box:key_step_prompt}
You are an expert system that gives you a question and a corresponding answer, please list in detail the key reasoning steps from the question to the answer, make sure that the reasoning steps are clear and complete, and include all possible solutions. You should pretend not to know the basic truth answer beforehand.

Input:
Question:

\{question\}

Answer:

\{answer\}

Output requirements:\\
1. Only include the essential key steps, and don't output unnecessary words\\
2. For each solution, record:\\
-logical\_conclusion: The set of each key step of the solution, from Step 1 all the way to the answer\\
3. A problem may contain more than one way of reasoning, so make sure you don't miss any possible solutions.\\
4. Important: Output only JSON array with no additional information.\\
5. Don't add useless words to the process\\

Here is the json output format:\\
\#\# Output format
\begin{verbatim}
[
{{
"solution1": {{
"logical_conclusion": ["step1:","step2:","step3:",...]
}}
}}
]
\end{verbatim}
\end{tcolorbox}

\refstepcounter{promptbox}
\begin{tcolorbox}[breakable, colback=gray!5!white, colframe=gray!75!black, 
title=Recall Evaluation Prompt, boxrule=0.5mm, width=\textwidth, arc=3mm, auto outer arc]
\label{box:recall_prompt}
\# Task Overview\\
You are an expert system for verifying solutions to text-based problems.    Your task is to match the ground truth middle steps with the provided solution.

\# INPUT FORMAT:\\
1. Problem: The original question/task\\
2. A Solution of a model\\
3. Ground Truth: Essential steps required for a correct answer\\

\# MATCHING PROCESS:\\
You need to evaluate each ground truth middle step against the solution, following these criteria:\\

\#\# Match Criteria:\\
- **Exact Match or Equivalent Logical Step**: A ground truth step is considered **Matched** if:\\
- It appears exactly in the solution **OR**\\
- The same logical reasoning or idea is clearly expressed, even if wording or format differs.\\
- **Numerical and Conceptual Consistency**: All key numbers, equations, or transformations should align conceptually with the ground truth.\\
- **Step-by-Step Evaluation**: Each ground truth step must be assessed individually and sequentially.\\
- **Final Answer Check**: Determine if the overall reasoning process leads to the correct final answer.\\

\# OUTPUT FORMAT:
\begin{verbatim}
[
  {{
    "step_index": <integer>,
    "judgment": "Matched" | "Unmatched",
    "correct_answer": "true" | "false"
  }}
]
\end{verbatim}

\# ADDITIONAL RULES:\\
1. **Strict JSON Output**: Output only the JSON array with no additional text or explanations.\\
2. **No Omitted Steps**: Every step in `Ground Truth` must be evaluated.\\

\# EDGE CASE HANDLING:\\
- If a step is conceptually equivalent but reworded, it is still considered **Matched**.\\
- If numerical transformations are equivalent (e.g., same formula in a different form), it is **Matched**.\\
- If the reasoning process does not lead to the correct final answer, \texttt{"correct\_answer": "false"}.\\

Here is the problem, answer, solution, and the ground truth middle steps:\\

[Problem]

\{question\}

[Answer]

\{answer\}

[Solution]

\{solution\}

[Ground Truth Information]

\{gt\_annotation\}
\end{tcolorbox}

\refstepcounter{promptbox}
\begin{tcolorbox}[breakable, colback=gray!5!white, colframe=gray!75!black, 
title=Precision Evaluation Prompt, boxrule=0.5mm, width=\textwidth, arc=3mm, auto outer arc]
\label{box:precision_prompt}
\# Task Overview\\
Given a solution with multiple reasoning steps for a text problem, reformat it into well-structured steps and evaluate their correctness.

\# Step 1: Reformatting the Solution\\
Convert the unstructured solution into distinct reasoning steps while:\\
- Preserving all original content and order\\
- Not adding new interpretations\\
- Not omitting any steps\\

\#\# Step Types\\
1. Logical Inference Steps\\
   - Contains exactly one logical deduction\\
   - Must produce a new derived conclusion\\
   - Cannot be just a summary or observation\\

2. Background Information Steps\\
   - External knowledge or question context\\
   - No inference process involved\\

\#\# Step Requirements\\
- Each step must be atomic (one conclusion per step)\\
- No content duplication across steps\\
- Initial analysis counts as background information\\
- Final answer determination counts as logical inference\\

\# Step 2: Evaluating Correctness\\
Evaluate each step against:\\

\#\# Ground Truth Matching\\
For logical inferences:\\
- Conclusion must EXACTLY match or be DIRECTLY entailed by ground truth\\

\#\# Reasonableness Check (if no direct match)\\
Step must:\\
- Premises must not contradict any ground truth or correct answer\\
- Logic is valid\\
- Conclusion must not contradict any ground truth \\
- Conclusion must support or be neutral to correct answer\\

\#\# Judgement Categories\\
- "Match": Aligns with ground truth\\
- "Reasonable": Valid but not in ground truth\\
- "Wrong": Invalid or contradictory\\
- "N/A": For background information steps\\

\# Output Requirements\\
1. The output format MUST be in valid JSON format without ANY other content.\\
2. For highly repetitive patterns, output it as a single step.\\
3. Output maximum 35 steps. Always include the final step that contains the answer.\\
4. correct\_answer: Whether the whole reasoning process produces the right answer.\\

Here is the json output format:\\
\#\# Output Format\\
\begin{verbatim}
[
  {{
    "step_type": "logical inference|background information",
    "premise": "Evidence",
    "conclusion": "Step result",
    "judgment": "Match|Reasonable|Wrong|N/A"
    "correct_answer": "true|false"
  }}
]
\end{verbatim}

Here is the problem, and the solution that needs to be reformatted to steps:\\

[Problem]

\{question\}

[Solution]

\{solution\}

[Correct Answer]

\{answer\}

[Ground Truth Information]

\{gt\_annotation\}
\end{tcolorbox}

\refstepcounter{promptbox}
\begin{tcolorbox}[breakable, colback=gray!5!white, colframe=gray!75!black, 
title=Model Output Reformat Prompt, boxrule=0.5mm, width=\textwidth, arc=3mm, auto outer arc]
\label{box:reformat_prompt}
I will present you with a solution to a problem. Unfortunately, the solution lacks proper paragraphing, making it hard to read. Your task is to improve readability by reformatting the solution into well-structured paragraphs. Follow these specific guidelines:

* Insert \verb|\n\n|  for paragraph breaks within the original solution. Do **NOT** alter any content of the original solution (the only exception is for itemized lists; see below).\\
  - Each paragraph should represent a distinct, concise reasoning step that logically advances the solution.\\
  - Reasoning steps can include case discussions, formula simplifications, or formula derivations. Each of these should be treated as an individual reasoning step and paragraphed accordingly.\\
  - If an introductory analysis exists in the original solution, treat it as an initial reasoning step and place it as the first paragraph.\\
  - Do **NOT** place any formulas in their own separate paragraphs; instead, include them within the same paragraph as the preceding text to form a cohesive reasoning step.\\

* For any itemized lists (ordered or unordered), convert them into a written format, such as "First/Second/Third." This is the **ONLY** content modification allowed.

* Avoid making paragraphs too lengthy, as long paragraphs might contain multiple reasoning steps that should be paragraphed separately.

* Disregard the accuracy of the solution content. Do **NOT** alter any of the original solution's content; focus solely on structuring it into logical, readable paragraphs.

* Reply with the reformatted solution directly.

--------------------------------------------------

Here is the problem, and the solution that needs to be reformatted:

[Problem]

\{problem\}

[Solution]

\{response\}
\end{tcolorbox}

\refstepcounter{promptbox}
\begin{tcolorbox}[breakable, colback=gray!5!white, colframe=gray!75!black, 
title=First Correct Answer Extraction Prompt, boxrule=0.5mm, width=\textwidth, arc=3mm, auto outer arc]
\label{box:correct_answer_prompt}
The following is a problem and a solution (split into paragraphs, enclosed with tags and indexed from 0):

[Problem]

\{problem\}

[Correct Answer]

\{answer\}

[Solution]

\{tagged\_response\}

Your task is to review and critique the solution paragraph by paragraph. Once you identify an correct answer in a paragraph, return the index of the paragraph where the earliest correct answer occurs. Otherwise, return the index of -1 (which typically denotes "not found").

Please put your final answer (i.e., the index) in \\boxed{{}}.
\end{tcolorbox}

\refstepcounter{promptbox}
\begin{tcolorbox}[breakable, colback=gray!5!white, colframe=gray!75!black, 
title=Reflection Quality Prompt, boxrule=0.5mm, width=\textwidth, arc=3mm, auto outer arc]
\label{box:reflection_quality}
Here's a refined prompt that improves clarity and structure:

\# Task
Evaluate reflection steps in a problem-solving solutions, where reflections are self-corrections or reconsiderations of previous statements.

\# Reflection Step Identification \\
Reflections typically begin with phrases like:\\
- "But xxx"\\
- "Alternatively, xxx" \\
- "Maybe I should"\\
- "Let me double-check"\\
- "Wait xxx"\\
- "Perhaps xxx"\\
It will throw an doubt of its previously reached conclusion or raise a new thought.

\# Evaluation Criteria\\
Correct reflections must:\\
1. Reach accurate conclusions aligned with ground truth\\
2. Use new insights to find the mistake of the previous conclusion or verify its correctness. \\

Invalid reflections include:\\
1. Repetition - Restating previous content or method without new insights\\
2. Wrong Conclusion - Reaching incorrect conclusions vs ground truth\\
3. Incompleteness - Proposing but not executing new analysis methods\\
4. Other - Additional error types\\

\# Input Format

[Problem]

\{question\}

[Think Content]

\{think\_content\}

[Ground Truth]

\{gt\_annotation\}

\# Output Requirements\\
1. The output format must be in valid JSON format without any other content.\\
2. Output maximum 30 reflection steps.\\

Here is the json output format:\\
\#\# Output Format
\begin{verbatim}
[
  {{
    "conclusion": "One-sentence summary of reflection outcome",
    "judgment": "Correct|Wrong",
    "error_type": "N/A|Repetition|Wrong Conclusion|Incompleteness|Other"
  }}
]
\end{verbatim}

\# Rules
1. Preserve original content and order\\
2. No new interpretations\\
3. Include ALL reflection steps\\
4. Empty list if no reflections found\\
5. Direct JSON output without any other output\\
\end{tcolorbox}

\end{document}